\documentclass[letterpaper,10pt,conference]{ieeeconf}  

\IEEEoverridecommandlockouts                             
\PassOptionsToPackage{export}{adjustbox}
\usepackage{definitions}
\usepackage[T1]{fontenc}
\usepackage{tikz}
\usepackage{flushend}

\usepackage{dsfont}
\usepackage[cal=pxtx,scr=boondox]{mathalpha}
\usepackage{bm}
\usepackage{yhmath}
\usepackage{adjustbox}
\usepackage{mathtools}
\usepackage{amsmath}
\usepackage{amssymb} 
\usepackage[linesnumbered,ruled,vlined]{algorithm2e}
\usepackage{url}
\usepackage{graphics}
\usepackage{graphicx}
\usepackage{epsfig}
\usepackage{textcomp}
\usepackage{cite}
\usepackage{bm}
\usepackage{subcaption}
\usepackage[labelfont=bf]{caption}
\usepackage{booktabs}
\usepackage[english]{babel}
\usepackage[utf8]{inputenc}
\usepackage{theorem}
\usepackage{pifont}
\usepackage{cuted}
\usepackage[dvipsnames]{colortbl,xcolor}
\usepackage{multirow}
\usepackage{siunitx}
\usepackage{gensymb}
\usepackage{stmaryrd}

\let\labelindent\relax
\usepackage{enumitem}

\newcommand{\ttx}{\texttt{x}}
\newcommand{\tty}{\texttt{y}}
\newcommand{\ttz}{\texttt{z}}

\newcommand{\tcmd}{\text{cmd}}

\newcommand{\scatter}{\ttt{Scatter\,}}
\newcommand{\maze}{\ttt{Maze\,}}

\newcommand{\tc}{\text{TC}}

\newcommand{\ind}{\mathds{1}}
\newcommand{\plc}[1]{{{\pi}}_\ttt{#1}}
\newcommand{\best}{\cellcolor{blue!15}}
\newcommand{\bess}{\cellcolor{ForestGreen!15}}

\title{\LARGE \bf REASAN: Learning Reactive Safe Navigation for Legged Robots}

 \author{Qihao~Yuan, Ziyu~Cao, Ming~Cao, and Kailai~Li%
	 	\thanks{Qihao Yuan and Kailai Li are with the Bernoulli Institute for Mathematics, Computer Science and Artificial Intelligence, University of Groningen, The Netherlands. Ziyu Cao is with the Department of Electrical Engineering, Linköping University, Sweden. Ming Cao is with the Engineering and Technology Institute Groningen, University of Groningen, The Netherlands.
		 		E-mails: \tt{qihao.yuan@rug.nl, ziyu.cao@liu.se, m.cao@rug.nl, kailai.li@rug.nl}.}%
	 	\thanks{\textit{(Corresponding author: Kailai Li.)}}%
	 }

\begin{document}
	
	\maketitle
	\thispagestyle{empty}
	\pagestyle{empty}
	
	\begin{abstract}
		We present a novel modularized end-to-end framework for legged reactive navigation in complex dynamic environments using a single light detection and ranging (LiDAR) sensor. The system comprises four simulation-trained modules: three reinforcement-learning (RL) policies for locomotion, safety shielding, and navigation, and a transformer-based exteroceptive estimator that processes raw point-cloud inputs. This modular decomposition of complex legged motor-control tasks enables lightweight neural networks with simple architectures, trained using standard RL practices with targeted reward shaping and curriculum design, without reliance on heuristics or sophisticated policy-switching mechanisms. We conduct comprehensive ablations to validate our design choices and demonstrate improved robustness compared to existing approaches in challenging navigation tasks. The resulting reactive safe navigation (REASAN) system achieves fully onboard and real-time reactive navigation across both single- and multi-robot settings in complex environments. We release our training and deployment code at \ttt{https://github.com/ASIG-X/REASAN}
	\end{abstract}
	
	
	\section{Introduction}\label{sec:intro}
	Legged robots offer distinct advantages given their universal mobility, with expanding application scenarios ranging over search and rescue, logistics, entertainment, industrial inspection, and forestry inventories~\cite{frey2025boxi,arnold2025leva,ma2025learning,mattamala2025building}. Recent advances in quadrupedal locomotion have demonstrated remarkable performance, particularly, in handling complex static terrains~\cite{chen2025learning,hoeller2024anymal,agarwal2023legged}. However, legged navigation in everyday, human-centric environments remains fundamentally challenging due to inherent complexities in both methodology and engineering practices, including the need for high-performance locomotion, handling dynamically changing scenes and moving obstacles, sensor integration, and operating under limited onboard resources~\cite{scheidemann2025obstacle}. 
	
	Safe navigation for mobile robots has traditionally focused on collision-free path planning using methods based on sampling, search, and optimization, which typically require a prior map of the environment to generate feasible trajectories~\cite{gaertner2021collision,vernaza2009search,liao2023walking}. However, such pipelines can pose significant challenges for online deployment in dynamic environments, where runtime efficiency is critical and timely mapping of the surroundings becomes increasingly intractable~\cite{reijgwart2024waverider,scheidemann2025obstacle}. An alternative is to enable reactive behaviors that adapt directly to changing ambient conditions through single-step actions. This strategy allows robots to respond rapidly to instantaneous perceptual variations and is particularly appealing when paired with agile legged locomotion capabilities~\cite{wang2025omni}. 
	\begin{figure}
		\vspace{2mm}
		\centering
		\includegraphics[width=0.97\linewidth]{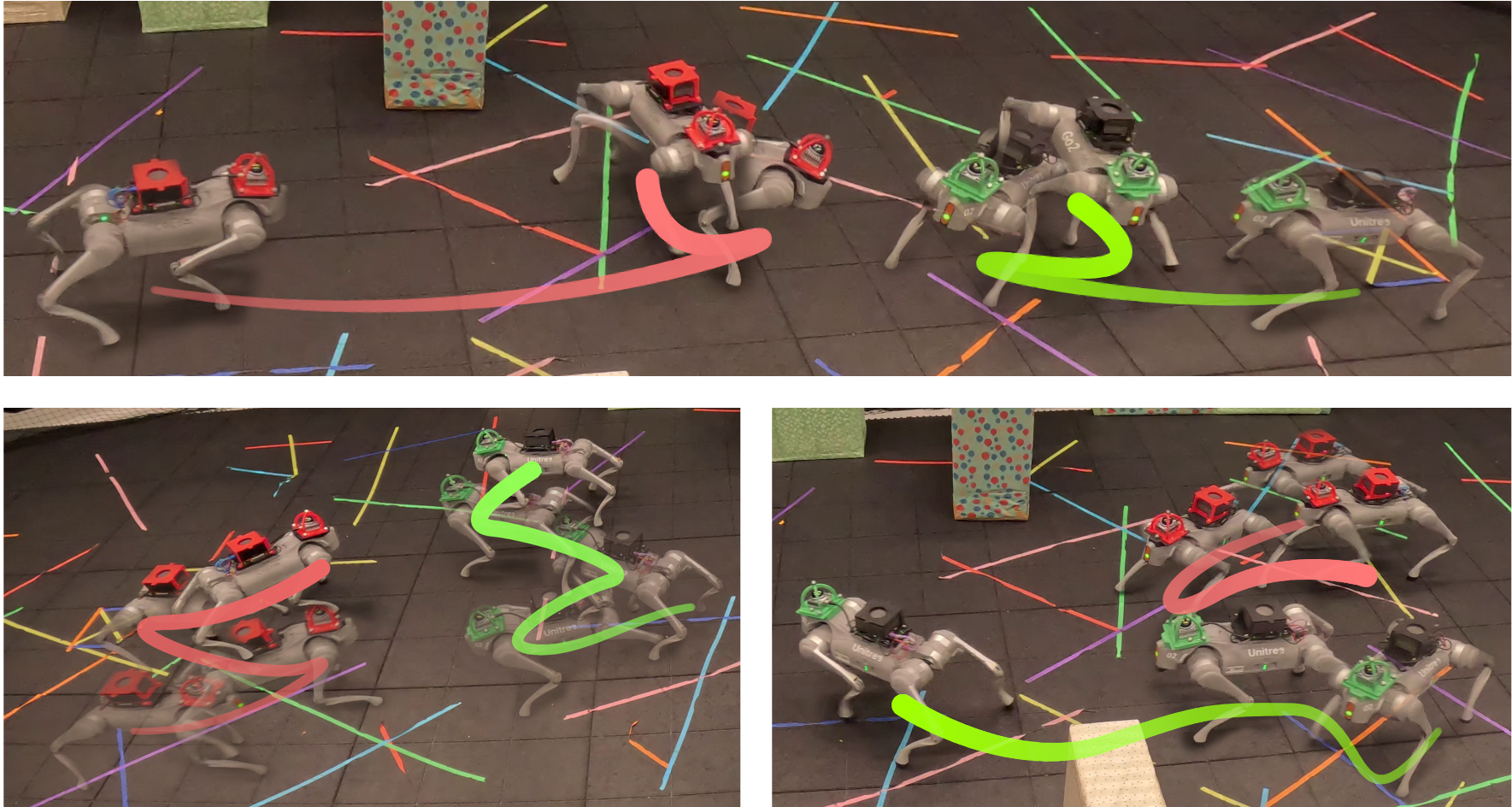}
		\caption{REASAN deployed onboard two quadrupedal robots.}
		\label{fig:front}
		\vspace{-2mm}
	\end{figure}
	
	Considerable efforts have been dedicated to achieving reactive obstacle avoidance for legged navigation. Beyond popular methodologies such as model predictive control~\cite{gaertner2021collision}, one promising strategy is to directly generate high-order motions, e.g., accelerations, reactively to the robot state and (local) map of the environment. Building on the formulation of Riemannian Motion Policies (RMPs)~\cite{ratliff2018riemannian}, a reactive local planner was proposed in \cite{reijgwart2024waverider} using a multi-resolution volumetric map for parallelizable obstacle avoidance onboard unmanned aerial vehicles. This framework was later adapted in~\cite{scheidemann2025obstacle} to reactively avoid local and dynamic obstacles for leader following using quadrupedal robots. In~\cite{mattamala2022efficient}, an RMP-based reactive controller was developed in combination with signed/geodesic distance functions computed from 2.5D elevation maps for quadrupedal visual teach and repeat tasks. These systems exhibit improved onboard deployability in complex environments; however, due to the reliance on intermediate representations (such as vector fields) and optimization-based solutions, the reported overall agility of legged robots remains limited when reacting to dynamic obstacles, often resulting in slow walking speeds~\cite{scheidemann2025obstacle}.
	
	Learning-based methods have also shown strong promise for enabling reactive behavior for legged robots in dynamic environments~\cite{schwarke2025rsl}. \cite{seo2022learning} introduced a hierarchical framework that combines a high-level navigation policy learned from human demonstrations via imitation learning with a low-level gait controller trained using reinforcement learning (RL). In~\cite{he2024agile}, a hierarchical control framework, \textit{Agile But Safe} (ABS), was proposed to enable fast and collision-free quadrupedal locomotion using depth images. An RL-based agile policy is trained for rapid goal reaching amid simple obstacles given exteroceptive input from a ray prediction network. However, a policy-conditioned reach–avoid (RA) value network is required to evaluate collision risk and prevent failures, triggering an additional recovery policy that tracks a 2D twist command computed through a constrained optimization problem to reduce the RA value. As a result, the system exhibits only relatively simple reactive behavior, primarily handling obstacles that block straight-line egomotion. Moreover, reliance on RGB-D sensing limits its ability to perceive omnidirectional obstacles and to operate reliably in low-light environments.
	
	In~\cite{wang2025omni}, an RL-based omnidirectional obstacle avoidance system was introduced for quadrupedal robots using LiDAR perception while tracking velocity commands. Although the system is designed end-to-end, an intermediate avoidance velocity command is explicitly computed from perceived nearby obstacles and used as a heuristic reference during training. Consequently, the robot can primarily handle dynamic objects only in relatively sparse environments while moving at moderate speeds (approximately \SI{1}{\meter/\second}). Moreover, the policy was not fully deployed onboard but rather through a cabled connection for velocity tracking, limiting its applicability to real-world navigation tasks~\cite{chen2025learning}.
	
	The aforementioned state-of-the-art approaches primarily focus on enabling isolated capabilities (e.g., high-speed or omnidirectional avoidance). Moreover, these methods often depend on explicitly generated intermediate twist commands, either through heuristics~\cite{wang2025omni} or more sophisticated control-theoretic methods~\cite{he2024agile}. Overall, there remains a lack of holistic and systematic methodologies as well as engineering practices towards achieving robust reactive navigation fully onboard legged robots.
	
	\subsection*{Contributions}
	We present REASAN (\textbf{Rea}ctive \textbf{Sa}fe \textbf{N}avigation), a novel modularized end-to-end framework for learning LiDAR-based reactive legged navigation in complex environments. Our system comprises three policy networks for locomotion, safety-shield, and navigation, and an exteroceptive estimator. 
	\begin{itemize}[leftmargin=*]
		\item Each policy uses a lightweight network, trained in simulation using standard RL with tailored reward shaping and curriculum design without relying on intermediate heuristics or sophisticated policy-switching mechanisms.
		\item A new Transformer-based network is introduced for estimating ray-based exteroceptive representation, trained via supervised learning in simulation, enabling dynamic obstacle awareness directly from raw LiDAR scans.
		\item Ablations in simulation validate our modularized design, showing improved robustness in complex navigation tasks compared with existing monolithic approaches. Extensive real-world experiments demonstrate fully onboard and real-time reactive navigation of legged robots in both single- and multi-agent settings across complex environments, including detouring and escaping dead ends.
		\item We open-source our complete training and deployment implementation, including a customized IsaacLab raycaster that supports an arbitrary number of dynamic objects.
	\end{itemize}
	\begin{figure*}
		\vspace{1mm}
		\centering
		\includegraphics[width=\textwidth]{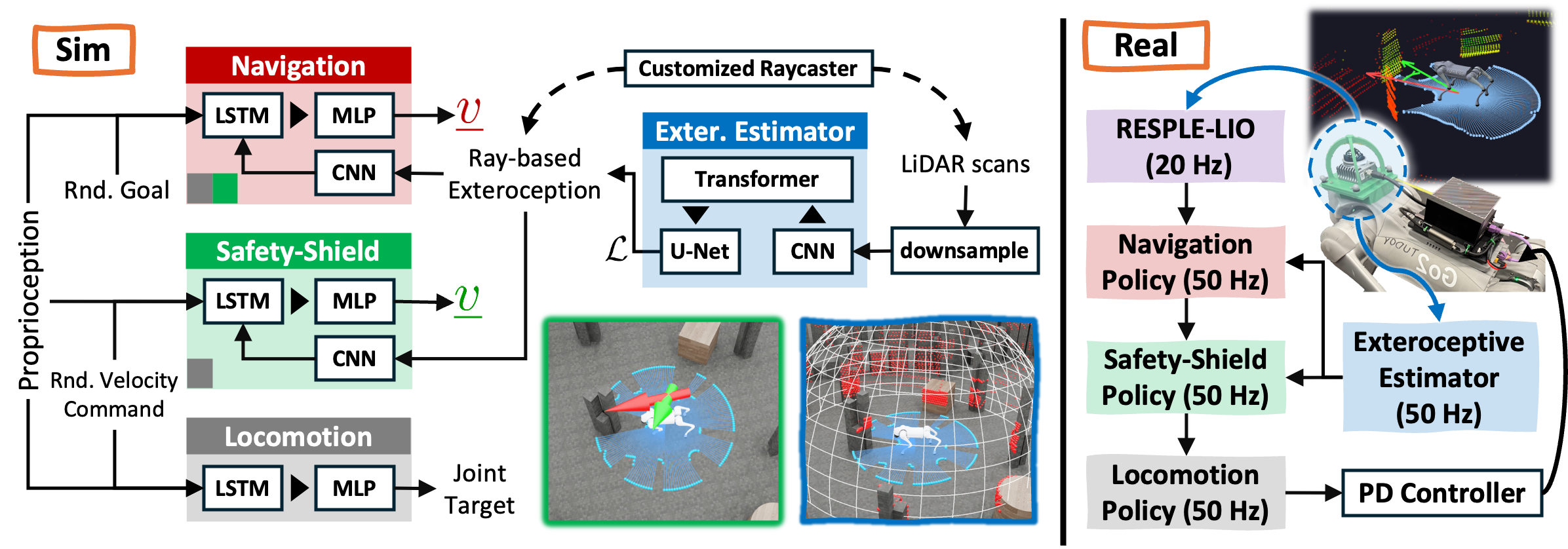}
		\caption{System pipeline of REASAN with a highly modularized design for both training and deployment.}
		\label{fig:pipeline}
		\vspace{-4mm}
	\end{figure*}
	
	\section{System Pipeline}\label{sec:system_pipeline}
	As illustrated in \figref{fig:pipeline}, our proposed framework is highly modularized for both training and deployment. The locomotion policy network $\plc{loco}$ tracks arbitrary 2-d velocity commands in the base frame and outputs the 12-d joint target positions executed by the internal PD controller for motor control. It takes the proprioception including the 12-d joint positions and velocities, the base angular velocity and projected gravity in the base frame, as well as the previous action. The safety-shield policy $\plc{safe}$ transforms an arbitrary velocity command into a safe one for locomotion, enabling reactive obstacle avoidance during velocity tracking. Given a goal-reaching target, the navigation policy $\plc{nav}$ generates velocity commands for map-free guidance, enabling more complex reactive behaviors such as detouring around obstacles or escaping dead ends. Both navigation and safety-shield policies take the same exteroception represented by $180$ equidistant rays covering a \SI{360}{\degree} FoV, which are predicted by the exteroceptive estimator from raw LiDAR scans.
	
	We first train the locomotion policy, then freeze its parameters and train the safety-shield policy on top of it. The navigation policy is further trained while running inference through the safety-shield and locomotion policies. All three policies are trained sequentially in IsaacLab~\cite{mittal2023orbit} using the PPO algorithm~\cite{schulman2017proximal} implemented in the RSL-RL framework~\cite{schwarke2025rsl}. Each policy network adopts the simple architecture, all consisting of a single-layer LSTM (256 hidden units) followed by a three-layer MLP (512-256-128 hidden dimensions). To interface with the predicted rays, we incorporate a 1-d CNN of 64-d output latent embedding in both safety-shield and navigation policies. The exteroceptive estimator is Transformer-based~\cite{vaswani2017attention} and trained through supervised learning with data collected in simulation while executing the safety-shield policy. The navigation policy is deliberately excluded during data collection to prevent goal-reaching bias, ensuring that the exteroceptive estimator remains general for both the safety-shield and navigation policies. This design also enables fully separate training of the navigation policy and the exteroceptive estimator. 
	
	During deployment, we integrate the four modules sequentially, enabling the full pipeline from exteroceptive estimation to motor control. A modified version of RESPLE~\cite{cao2025resple} is exploited for efficient LiDAR-inertial localization to provide goal position for navigation.
	
	\section{Modularized Learning}\label{sec:learning}
	We now introduce the proposed modularized learning framework aligned with the four modules shown in \figref{fig:pipeline}. The key aspects of each module are outlined below; more details are available in our open-source implementation.
	
	\subsection{Locomotion Policy}\label{subsec:loco}
	We train a robust locomotion policy to track arbitrary velocity commands $(v_\ttx,v_\tty,\omega_\ttz)$, representing two-dimensional linear and angular velocity in the base frame, up to \SI{2.5}{\meter/\second}, \SI{1.5}{\meter/\second}, and \SI{3}{\radian/\second}, respectively. The reward functions and curriculum learning follow established practices of learning blind locomotion with velocity tracking~\cite{chen2025slr}. 
	\begin{figure}[t!]
		\vspace{1mm}
		\centering
		\setlength{\tabcolsep}{1pt} 
		\begin{tabular}{ccc}
			\includegraphics[width=0.16\textwidth]{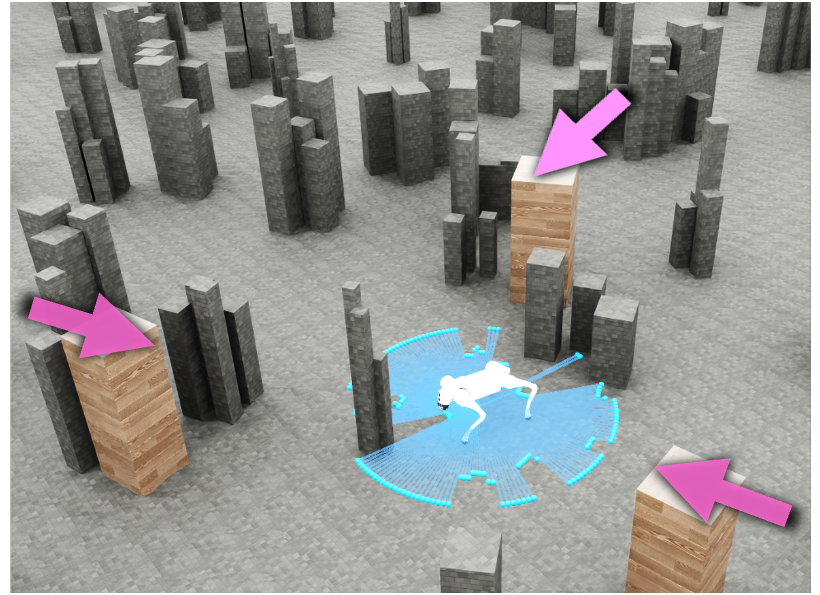}&
			\includegraphics[width=0.157\textwidth]{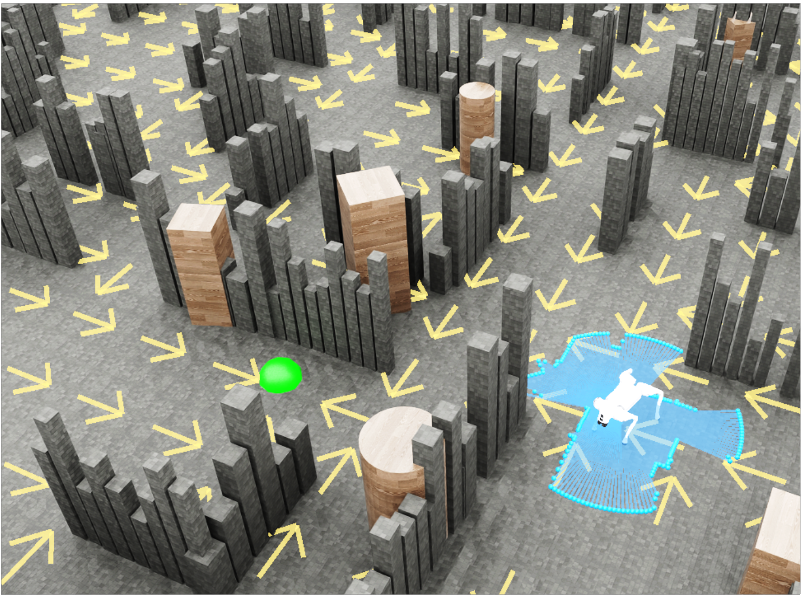} &
			\includegraphics[width=0.157\textwidth]{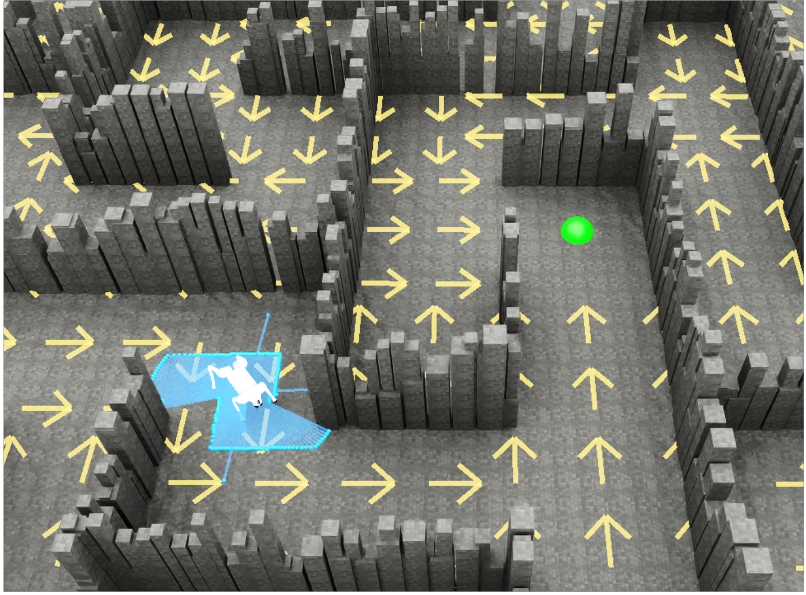}\\
			{(a) \ttt{SoftChase} \ttt{}} &{(b) \ttt{Scatter}} &{(c) \ttt{Maze}}\\
		\end{tabular}
		\caption{Environments for learning the safety-shield (a) and navigation policies (b)-(c). Yellow arrows indicate guidance.}
		\label{fig:simEnv}
		\vspace{-3mm}
	\end{figure}

	\subsection{Safety-Shield Policy}\label{subsec:shield}
	Reactive obstacle avoidance is solely enabled by the safety-shield policy, which transforms the high-level navigation velocity command into a safe one for locomotion given the $180$-ray exteroception. The rationale for isolating reactive obstacle avoidance from low-level motor control is to reduce the difficulty of learning effective policies from complex perceptual inputs. Given our practice, the presence of dynamic obstacles can introduce highly unstable and noisy reward signals, making it challenging for the robot to directly acquire reliable collision-free locomotion skills. This design choice also simplifies the policy network architecture: the same LSTM and MLP structures used for locomotion are retained, while a lightweight 1D-CNN is employed ahead to extract features from the ray-based exteroceptive representation, treated as a single-channel 1D image. In addition, it enables direct end-to-end reactive avoidance without relying on intermediate twist commands generated by heuristics or optimization, nor on sophisticated mechanism such as policy switching, in contrast to prior works~\cite{wang2025omni,he2024agile}. This strategy further simplifies reward shaping by cleanly decoupling from other design considerations.
	
	\subsubsection*{Reward functions}
	Given an arbitrary velocity command $(v_{\ttx}^\tcmd,v_{\tty}^\tcmd,\omega_{\ttz}^\tcmd)$ in the base frame, we tune the following reward terms for tracking it with safety shield
	\begin{equation*}
		r_\text{track} = 4\cdot\exp\,(-4\Vert\uv^\tcmd - \uv\Vert^2) + 3\cdot\exp\,(-2(\omega_\ttz^\tcmd-\omega_\ttz)^2)\,,
	\end{equation*}
	where $\uv=[v_\ttx, v_\tty]^\top$ and $\omega_\ttz$ denote the robot's current velocity in the base frame. The following penalty is further applied to encourage smooth output commands
	\begin{equation*}
		r_\text{smooth}= -0.01 \cdot \Vert \ua_t - 2\ua_{t-1} + \ua_{t-2}\Vert^2\,,\eqwith
	\end{equation*}
	$\ua_t=[v_\ttx,v_\tty,\omega_\ttz]^\top$ representing the action taken at time step $t$. We penalize collision on body parts as follows
	\begin{equation*}
		r_\text{collision} = \textstyle -100\cdot\sum_{\iota}\ind\big\{||\underline{\rho}_\iota|| > 0.1\big\}\,,\eqwith\iota\in\mathcal{B}_\text{undesired}
	\end{equation*}
	being the body part where collisions are undesired, i.e., everywhere except the feet. $\underline{\rho}_\iota$ denotes the contact force on body part $\iota$, and $\ind$ is the indicator function. In addition, we penalize undesired velocity direction according to
	\begin{equation*}
		r_\text{vel} = -\exp(-2\cdot t^\tc)\,,\eqwith
	\end{equation*}
	$t^\tc$ denoting the time-to-collision along the exteroceptive ray that is closest to the robot's velocity direction. Along the $i$-th ray with measured distance $r_i$, it is computed according to  
	\begin{equation*}
		t_i^\tc= {r_i}/((\uv-\uv_i^\text{obstacle})^\top\uu_i)\,,\,i\in\{1,\cdots,180\}\,
	\end{equation*}
	where $\uu_i\in\Sbb^1\subset\R^2$ is the unit vector representing the direction of the $i$-th ray, and $\uv_i^\text{obstacle}\in\R^2$ denotes the velocity of the encountered obstacle. Furthermore, we penalize actions that exceed the limits of the target velocity of the locomotion policy as follows
	\begin{equation}\label{eq:limit}
		r_\text{limit}=-10\big((\vert v_\ttx\vert-2.5)_++(\vert v_\tty\vert-1.5)_++(\vert \omega_\ttz\vert-3)_+\big)\,,
	\end{equation}
	where function $(x)_+\coloneq\max\{x,0\}$\,.
	In practice, reactive avoidance behaviors may induce aggressive maneuvers that increase the risk of collision, particularly from the rear. Therefore, we penalize such over-reactive behavior with
	\begin{equation*}
		\begin{aligned}
			r_\text{over}= -5\cdot \ind\{\Vert\uv^\tcmd\Vert > 0.2,t_\tcmd^\tc > 2,\hat{\uv}^\top\hat{\uv}^\tcmd < -0.25\}\\
			-5 \cdot \ind\{\Vert\uv^\tcmd\Vert < 0.2,t_{\min}^\tc > 2.5,\Vert\uv\Vert> 0.2\}\,,
		\end{aligned}
	\end{equation*}
	with $\hat{\uv}$ and $\hat{\uv}^\tcmd$ being the unit vectors denoting the actual and commanded velocity directions, respectively. $t^\tc_\tcmd$ denotes the time-to-collision along the velocity command $\uv^\tcmd$, and $t_{\min}^\tc$ the minimum over all rays. The first term penalizes moving against a safe command; and the latter penalizes unnecessary motion when the commanded velocity is small.
	
	\subsubsection*{Curriculum design}
	The safety-shield policy is trained in two stages according to the terrain configuration. In Stage 1, we place scattered static obstacles and randomly sample input velocity commands within the limits of locomotion control. In Stage 2, we gradually introduce additional dynamic obstacles up to three as shown in \figref{fig:simEnv}-(a). In this setup, each dynamic obstacle first queries the robot’s current position as its goal and moves toward it at a random speed. Upon arrival, it updates the robot’s new position as the next goal, repeating this process. Because the obstacles do not chase the robot continuously but instead follow with a delay, this curriculum design introduces sufficient challenge while remaining tractable for learning an effective policy using PPO. To encourage richer robot-obstacle interactions, we additionally constrain the input velocity commands in half of the parallel environments (1024 in total) of the second stage to always head toward the origin, preventing the robot from simply escaping the dynamic obstacles.
	
	\subsection{Navigation Policy}\label{subsec:navigation}
	The navigation policy takes a goal position in the base frame as input and outputs a goal-reaching velocity command, given the ray-based exteroceptive observations. Learning a navigation policy in complex environments, such as those requiring long detours or escape from dead ends, is challenging due to PPO’s sample inefficiency and the sparsity of goal rewards, which often causes convergence to local optima. Therefore, besides the common goal-reaching reward structure, we generate an occupancy grid-based guiding direction field pointing toward the goal as privileged information using Dijkstra’s algorithm~\cite{johnson1973note}, as shown in \figref{fig:simEnv}-(b) and (c). The agent receives auxiliary rewards for following this guidance, which substantially simplifies exploration and accelerates learning in practice. The navigation policy can thus adopt a simple network architecture identical to that of the safety-shield policy, using a lightweight 1D-CNN followed by an LSTM and MLP to extract features from the exteroceptive observations.
	
	\subsubsection*{Reward functions}
	For goal-reaching, we set up a step-wise reward as follows
	\begin{equation}
		r_\text{goal} = 40\cdot \ind\{d < 0.5\} + 15\cdot \ind\{d< 1\}+5\cdot\ind\{d< 2\}\,,
	\end{equation}
	where $d$ is the current distance to the goal. To facilitate policy learning, we additionally reward progress of reaching  
	\begin{equation*}
		r_\text{progress} = 20\cdot\max\{d^\text{prv}- d,0\}\,,\eqwith
	\end{equation*}
	$d^\text{prv}$ being the previous distance to the goal. To follow the privileged optimal path, the following reward function is used
	\begin{equation}
		\begin{aligned}
			r_\text{guide} = \ind\{((\hat{\uv}^\tcmd)^\top\uu^\text{guide}) > 0.7, d\ge2\}+\ind\{d< 2\} \\
			-5\cdot\ind\{(\hat{\uv}^\tcmd)^\top{\uu}^\text{guide}) < 0.1, d \ge 2\}\,, 
		\end{aligned}
	\end{equation}
	where $\uu^\text{guide}\in\Sbb^1\subset\R^2$  is the current guiding direction. The first two terms encourage moving along the guidance and toward the goal, whereas the last term penalizes moving against the guidance when far from the goal. Additionally, we encourage fast goal-reaching by penalizing time consumption $r_\text{time} = -4\cdot\ind\{d \ge 2\} \cdot e^{-t_\text{rest}}$, with
	$t_\text{rest}$ being the remaining time in the current episode. To ensure compatibility and coordination with the safety-shield policy, we exploit the same action-penalty formulation in \eqref{eq:limit} for regulating the goal-reaching velocity command $\uv$.
	
	\subsubsection*{Curriculum design}
	We adopt a concise curriculum-learning strategy for the navigation policy, aligned with our goal of enabling basic reactive handling of complex scenarios and facilitating integration with more sophisticated downstream tasks, such as more strategic global planners. We perform safety-shield policy (\secref{subsec:shield}) inference throughout training, without requiring the navigation policy to learn additional obstacle-avoidance behaviors. Two types of training environments (512 each) are instantiated in parallel: \scatter and \maze, as shown in \figref{fig:simEnv}-(b) and (c), respectively. \scatter is set up with static obstacles, and we randomly spawn dynamic ones around the robot, gradually increasing their number up to 9 over the curriculum levels. The goal position is randomly sampled on the grids within a straight-line distance of \SI{7}{\meter}. In \maze, we design more complex navigation scenarios that require the policy to learn skills for handling detours or escaping dead ends. The goal is sampled within a straight-line of \SI{5}-\SI{7}{\meter} and a path of \SI{5}-\SI{15}{\meter} from origin to avoid overly complex detours.
	\begin{figure}[t!]
		\vspace{2mm}
		\centering
		\includegraphics[width=\linewidth]{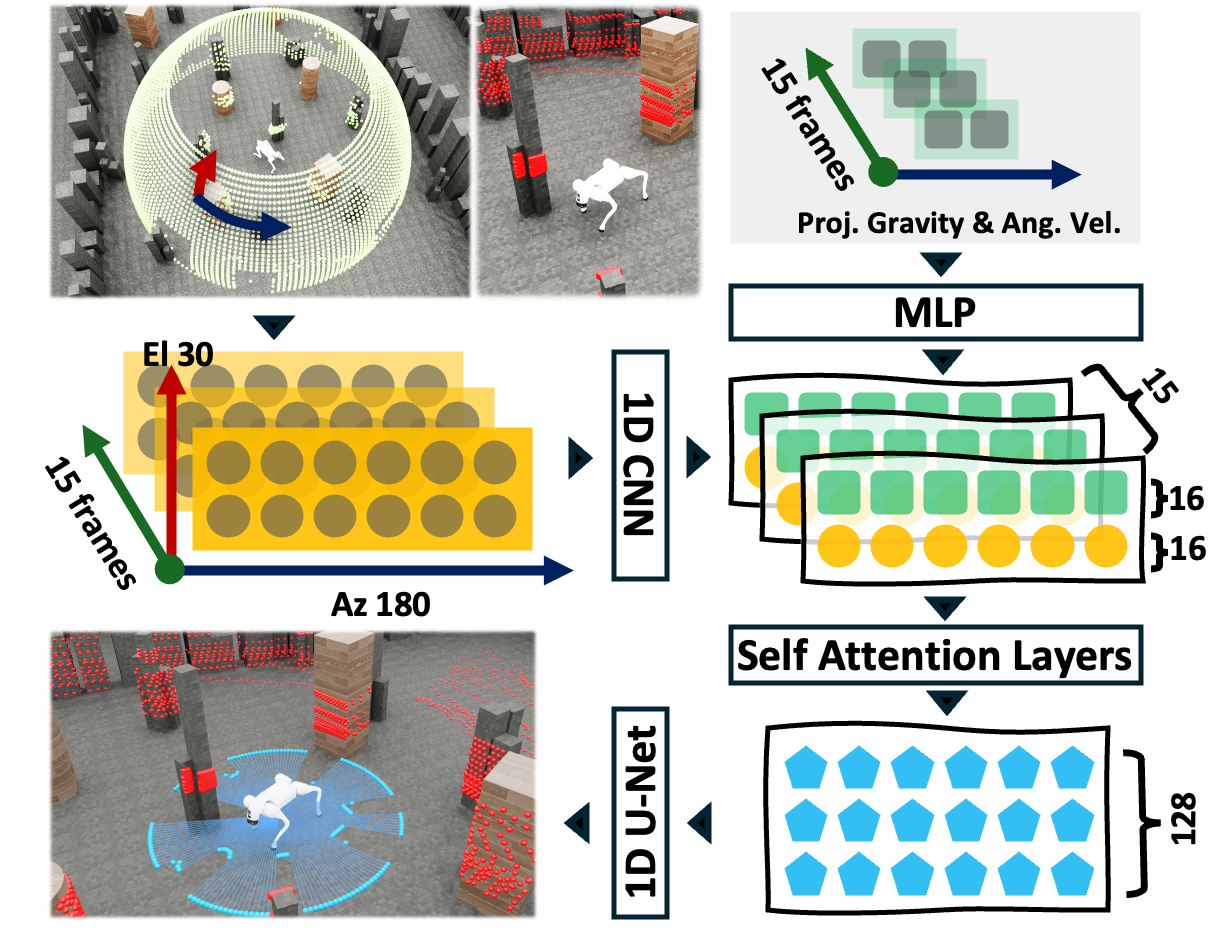}
		\caption{Network architecture of the exteroceptive estimator.}
		\label{fig:exter}
		\vspace{-4mm}
	\end{figure}
	
	\subsection{Exteroceptive Estimator}\label{subsec:estimator}
	As illustrated in \figref{fig:exter}, the proposed exteroceptive estimator follows a lightweight design. Each frame of points is first downsampled using a spherical grid with an angular resolution of \SI{2}{\degree} in both azimuth and elevation, retaining only the closest point in each cell. This yields a $30\times180$ circular depth image per frame with $\mathcal{O}(N)$ runtime complexity. We accumulate the past $15$ frames as input to a shared 1D-CNN, which compresses the channel dimension ($30\lsft{-5mu}\shortrightarrow\lsft{-5mu}16$) for each frame. In parallel, an MLP encodes the past $15$ proprioceptive measurements (projected gravity and angular velocity), producing a $16$-d feature for each frame. The LiDAR and proprioceptive features of all frames are then concatenated along the channel dimension at each spatial location to form tokens, which are subsequently fed into a Transformer encoder (with learnable temporal encodings applied to distinguish historical frames). Finally, the fused tokens are processed through a 1D U-Net~\cite{ronneberger2015u} to generate the final ray-based representation. The intuition for adopting a Transformer~\cite{vaswani2017attention} rather than sequential models such as GRUs is to better meet the runtime demands of exteroceptive estimation for timely collision awareness.
	
	\subsubsection*{Supervised learning}
	We adopt a Livox Mid-360 in IsaacLab for data collection in simulation, using rollout data generated by running the safety-shield policy while tracking random velocity commands. Note that the training pipeline is directly applicable to other LiDAR types given their respective scan patterns~\cite{wang2025omni}. A dataset of $5\times10^5$ items, each containing $15$ historical frames, is collected for training. Ground-truth ray distances are computed along horizontal directions relative to the ground using projected gravity. We propose a conservative MSE loss for training as follows
	\begin{equation}\label{eq:loss}
		\begin{aligned}
			\mL = 2 \cdot \mH(\hat{d}, d) \cdot \ind\{\hat{d} > d\} + \mH(\hat{d}, d) \cdot \ind\{\hat{d} \le d\} \\
			+ 0.05 \cdot \text{ReLU}(\hat{d} - d + 0.1) \cdot \ind\{d < 0.5\}\,,
		\end{aligned}
	\end{equation}
	where $\mH$ denotes the Huber loss, and $\hat{d}$ and $d$ represent the predicted and ground-truth ray distances, respectively, both clipped to \SI{3}{\meter} and normalized. The loss function discourages overestimation and intentionally introduces slight underestimation for distances below a threshold to enhance safety.
	
	\subsection{Implementation and Training Cost}
	As of the time of writing, the ray caster in IsaacLab does not support multiple objects, nor dynamic ones~\cite{mittal2023orbit}. We therefore extend the existing GPU-based ray caster to support an arbitrary number of dynamic objects, enabling fast training of the safety-shield and navigation policies as well as the exteroceptive estimator. To reduce the sim-to-real gap, we add a small amount of uniform noise to the distances returned by the ray caster.  All trainings are done on a single GeForce RTX 5090 GPU, with the individual costs profiled in Tab.~\ref{tab:cost}. Training the exteroceptive estimator takes about 10 hours and consumes 12 GB VRAM.
	\begin{table}[htbp]
		\centering
		\caption{Training cost for learning individual policies.}
		\setlength{\tabcolsep}{4pt} 
		\begin{tabular}{@{}l|ccccc@{}}
			\toprule
			\textbf{Policy module} & \textbf{\# Env.} & \textbf{Episode} & \textbf{\# Iter.} & \textbf{Time} & \textbf{VRAM}\\
			\midrule
			Locomotion 	  & $4096$ & \SI{20}{\second} & $10000$ & \SI{4}{\hour} & \SI{15}{\giga\byte} \\
			Safety-Shield & $1024$ & \SI{20}{\second} & $20000$ & \SI{15}{\hour} & \SI{12}{\giga\byte} \\
			Navigation    & $1024$ & \SI{15}{\second} & $10000$ & \SI{7}{\hour} & \SI{15}{\giga\byte} \\
			\bottomrule
		\end{tabular}
		\label{tab:cost}
		\vspace{-2mm}
	\end{table} 
	\begin{figure*}[t!]
		\vspace{2mm}
		\centering
		\setlength{\tabcolsep}{1pt}
		\begin{tabular}{ccccc}
			\includegraphics[width=0.195\textwidth]{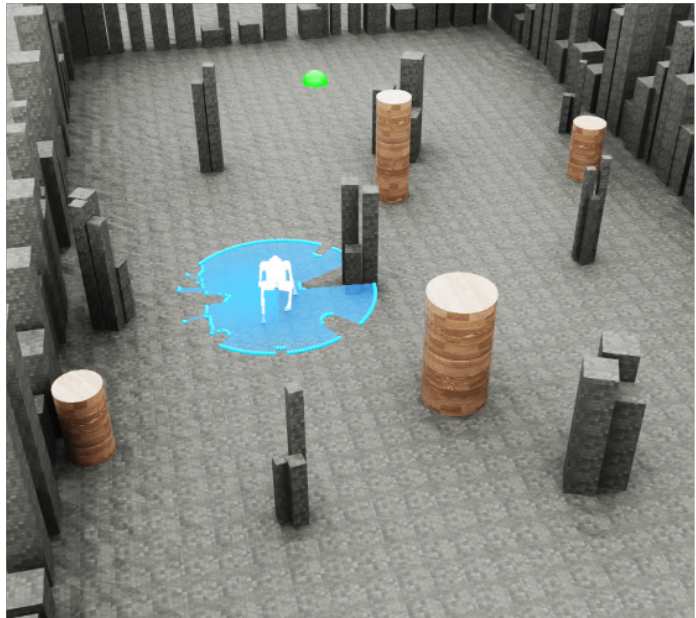}&
			\includegraphics[width=0.195\textwidth]{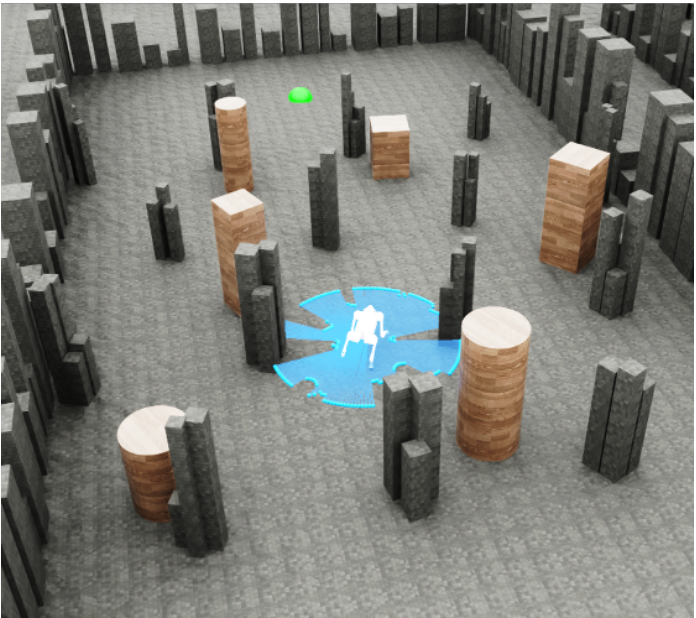}&
			\includegraphics[width=0.194\textwidth]{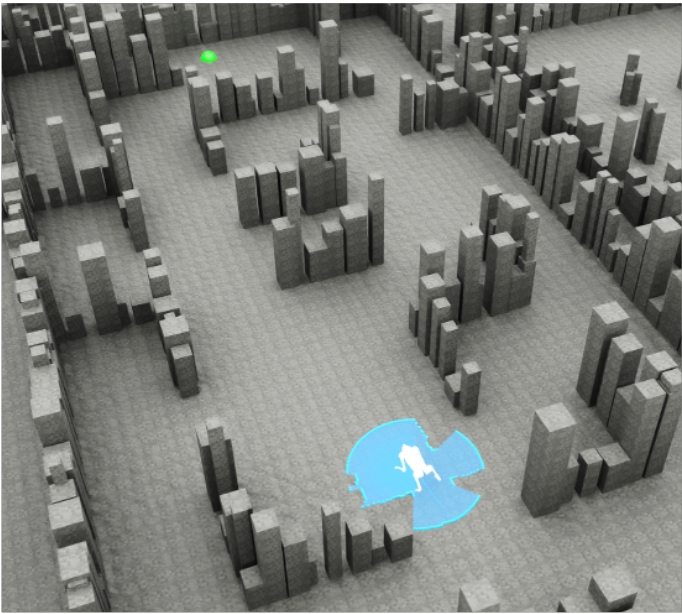}&
			\includegraphics[width=0.195\textwidth]{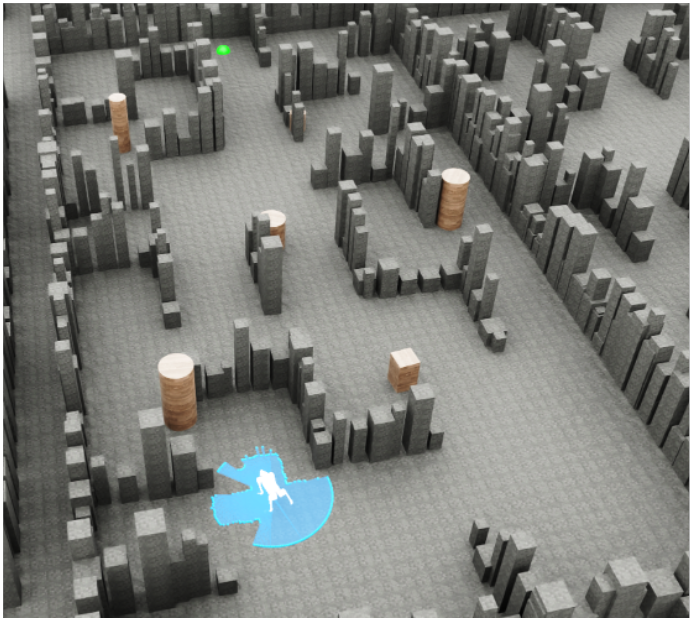}&
			\includegraphics[width=0.195\textwidth]{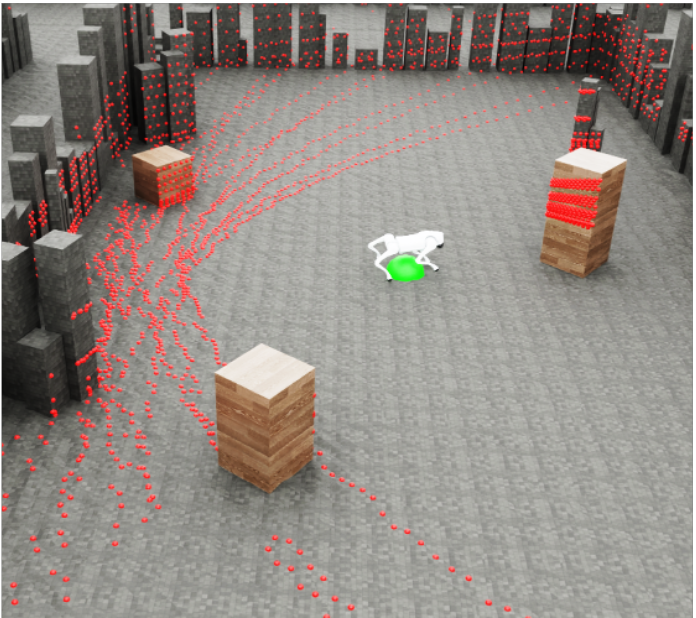}\\
			{(a) \ttt{ScaSparse}} &{(b) \ttt{ScaDense}} &{(c) \ttt{Maze}} &{(d) \ttt{DyMaze}} &{(e) \ttt{Hold}}\\
		\end{tabular}
		\caption{Evaluation scenarios in simulation. We simulate ray-based exteroception in (a)-(d) and raw LiDAR scans in (e).}
		\label{fig:isaac_eva}
		\vspace{-4mm}
	\end{figure*}
	
	\section{Evaluation in Simulation}\label{sec:ablation}
	\subsection{Modularization for Policy Learning}\label{subsec:eva_mod}
	We train a single end-to-end goal-reaching locomotion policy using the same network architecture and curriculum design as the safety-shield policy (\secref{subsec:shield}), following prior work~\cite{he2024agile}. During training, the end-to-end policy can successfully complete Stage~1 (static obstacles) but unsurprisingly fails to progress through Stage~2 (dynamic obstacles) due to the difficulty of training a compact monolithic network for achieving complex motor control. We conduct the following ablations on goal-reaching tasks across multiple simulation scenarios, with each setting evaluated over 10 rollouts.
	
	\subsubsection*{Obstacle avoidance}
	As shown in \figref{fig:isaac_eva}-(a) and (b), we set up two scenarios \ttt{ScaSparse} and \ttt{ScaDense}, containing scattered static and dynamic obstacles with large and small spacing, respectively, with \SI{15}{\second} per evaluation episode. In line with the end-to-end monolithic baseline, we train a safety-shield policy only up to Stage~1 to obtain basic \textbf{rea}ctive behavior (REA-stat). Besides the original safety-shield policy (REA), we also train a counterpart that explicitly rewards tracking an intermediate avoidance velocity computed heuristically according to prior work~\cite{wang2025omni} (REA-heu). As shown in Tab.~\ref{tab:mod}, we report success, termination, and timeout (TO) rates, with the best-performing policies for each scenario highlighted in green. The heuristic-based policy exhibits a conservative strategy that prioritizes maintaining large clearance from obstacles. Thus, it performs well in \ttt{ScaSparse}, however struggles significantly in denser environments due to frequent timeouts. In contrast, our proposed safety-shield policy (REA) and REASAN demonstrate substantially stronger adaptability across diverse dynamic conditions, achieving the best or near-best success rates.
	
	\subsubsection*{Navigation}
	Another two scenarios are shown in \figref{fig:isaac_eva}-(c) and (d). \ttt{Maze} has a static layout with dead ends and multiple feasible paths, while \ttt{DyMaze} adds dynamic obstacles that periodically block passages, each allocated \SI{30}{\second} per episode. Policies with the best performance are highlighted in blue in Tab.~\ref{tab:mod}. The standalone safety-shield system exhibits very limited capability in path finding (even worse than the baseline), but gains substantial improvement from the proposed navigation module, strongly supporting the effectiveness of modularization for reactive navigation in dynamic complex environments.
	\begin{table}[htbp]
		\vspace{-1mm}
		\centering
		\caption{Ablations for modularized policy learning.}
		\setlength{\tabcolsep}{4pt} 
		\begin{tabular}{c|l|rrr}
			\toprule
			& \textbf{System} & \textbf{Succ.} (\%) & \textbf{Term.} (\%) & \textbf{TO} (\%)\\
			\midrule
			\multirow{5}{*}{\rotatebox{90}{\ttt{{ScaSparse}}}} 
			& End2End & $77.8 \pm 3.8$ & $22.2 \pm 3.8$ & $0.0 \pm 0.0$\\
			& REA-stat & $79.3 \pm 4.0$ & $20.7 \pm 4.0$ & $0.0 \pm 0.0$ \\
			& \bess REA-heu  &\bess  $99.0 \pm 1.0$ &\bess $0.7 \pm 0.8$ &\bess $0.3 \pm 0.5$ \\
			& REA & $90.5 \pm 2.2$ & $9.5 \pm 2.2$ & $0.0 \pm 0.0$ \\
			& REASAN & $91.1 \pm 1.9$ & $8.9 \pm 1.9$ & $0.0 \pm 0.0$ \\
			\midrule
			\multirow{5}{*}{\rotatebox{90}{\ttt{{ScaDense}}}} 
			& End2End & $46.9 \pm 6.1$ & $52.8 \pm 6.3$ & $0.3 \pm 0.5$\\
			& REA-stat & $65.3 \pm 4.1$ & $34.7 \pm 4.1$ & $0.0  \pm 0.0$\\
			& REA-heu & $31.1 \pm 6.0$ & $6.6 \pm 2.2$ & $62.3 \pm 6.7$ \\
			& \bess REA & \bess $81.9 \pm 4.6$ &\bess $18.1 \pm 4.6$ &\bess $0.0 \pm 0.0$ \\
			& REASAN & $79.1 \pm 4.4$ & $20.3 \pm 4.3$ & $0.6 \pm 0.7$ \\
			\midrule
			\multirow{3}{*}{\rotatebox{90}{{\maze}}} 
			& End2End & $4.8 \pm 3.0$ & $33.5 \pm 5.7$ & $61.7 \pm 4.2$\\
			& REA    & $1.1 \pm 2.2$ & $1.4 \pm 2.7$ & $97.5 \pm 3.2$  \\
			&\best REASAN &\best $95.2 \pm 2.9$ &\best $1.9 \pm 2.6$ &\best $2.9 \pm 2.1$ \\
			\midrule
			\multirow{3}{*}{\rotatebox{90}{\scriptsize{\ttt{DyMaze}}}}
			& End2End & $6.8 \pm 3.1$ & $60.0 \pm 4.5$ & $33.2 \pm 3.3$ \\
			& REA & $1.2 \pm 2.0$ & $20.9 \pm 4.3$ & $77.9 \pm 4.4$ \\
			&\best REASAN & \best $68.2 \pm 2.3$ &\best $22.0 \pm 2.9$ &\best $9.8 \pm 3.3$ \\
			\bottomrule
		\end{tabular}
		\label{tab:mod}
		\vspace{-4mm}
	\end{table}
	
	\subsection{Exteroceptive Estimation}\label{subsec:exter}
	We evaluate three variants based on our original design and training pipeline in~\secref{subsec:estimator} for ablation: 1) REASAN-GRU: U-Net and Transformer are replaced by an MLP and a normal GRU, respectively; 2) REASAN-ConvGRU: Transformer replaced by a convolutional GRU~\cite{convlstm}; and 3) REASAN-Agg: The conservative loss~\eqref{eq:loss} is replaced by a single Huber loss. We set up a highly dynamic scenario, \ttt{Hold}, where the robot must safely maintain its starting position while avoiding $2$ -- $4$ obstacles moving along random paths intersecting at the starting position. Tab.~\ref{tab:exter} reports the collision-avoidance performance, inference time on a single GeForce RTX 5090 GPU (batch size 1), and the trained loss evaluated on a validation set of $5\times10^4$ samples. The proposed exteroceptive estimator achieves the highest success rate with superior inference quality and efficiency. REASAN-ConvGRU achieves a similar success rate, indicating that preserving spatial information is crucial for effective exteroception in dynamic scenarios, especially compared with the REASAN-GRU variant. However, it incurs a  $3.5\times$ inference cost compared with REASAN. Removing the conservative loss significantly degrades obstacle-avoidance performance, further justifying our loss-function design.
	\begin{table}[htbp]
		\centering
		\setlength{\tabcolsep}{4pt} 
		\caption{Ablations for exteroceptive estimator.}
		\begin{tabular}{l|rrrr}
			\toprule
			\textbf{System} & \textbf{Succ.} (\%) & \textbf{Term.} (\%)  &\textbf{Time} (\SI{}{ms}) &\textbf{Loss}\\
			\midrule
			REASAN-GRU        &$24.6 \pm 3.4$ &$75.4 \pm 3.4$ &$0.8 \pm 0.1$ &$0.0191$\\
			REASAN-ConvGRU    &$64.4 \pm 5.8$ &$35.6 \pm 5.8$ &$4.7 \pm 0.1$ &$0.0148$ \\
			REASAN-Agg   &$27.9 \pm 5.2$ &$72.1 \pm 5.2$ &$1.3 \pm 0.1$ &--\\
			\best REASAN &\best $65.8 \pm 3.4$ &\best $34.2 \pm 3.4$ &\best $1.3 \pm 0.1$ &\best $0.0135$ \\
			\bottomrule
		\end{tabular}
		\label{tab:exter}
	\end{table}
	
	\section{Real-World Experiments}\label{sec:experiment}
	\subsection{Hardware Setup and Deployment}\label{subsec:deploy}
	We deploy REASAN fully onboard a Unitree Go2 robot equipped with a Livox Mid-360 LiDAR and a Jetson AGX Orin (64 GB), as shown in \figref{fig:front} and \figref{fig:pipeline}. The trained neural networks are exported from PyTorch to ONNX format, using ONNX Runtime for fast onboard inference, with each module wrapped as a ROS2 node~\cite{onnxruntime,macenski2022robot}. To provide goal positions to the navigation module, we customize RESPLE by removing its global trajectory generation and mapping components, enabling fast LiDAR-inertial localization~\cite{cao2025resple}.
	\begin{figure}[t!]
		\centering
		\setlength{\tabcolsep}{1pt}
		\begin{tabular}{cc}
			\includegraphics[width=0.235\textwidth]{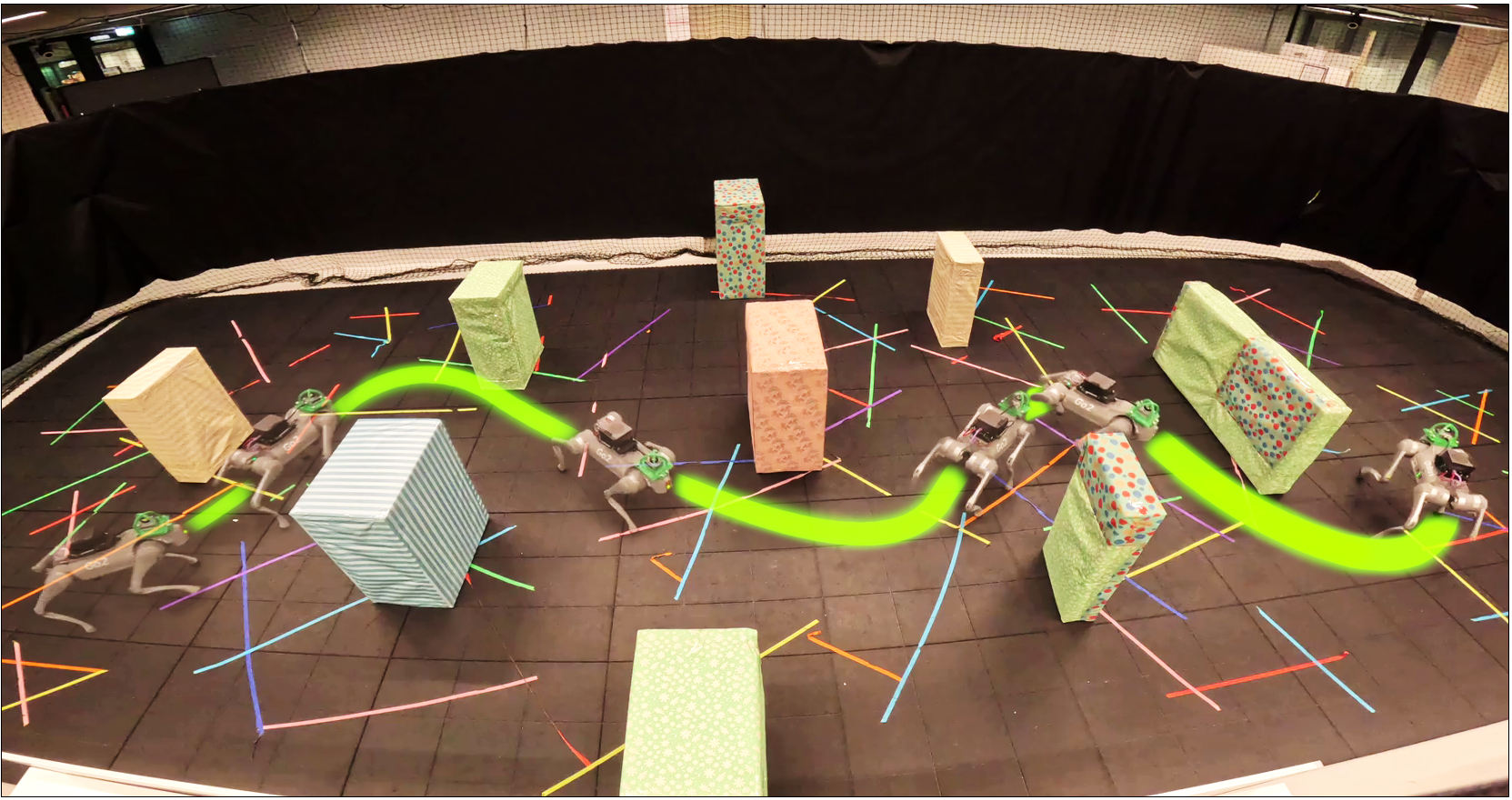}&
			\includegraphics[width=0.235\textwidth]{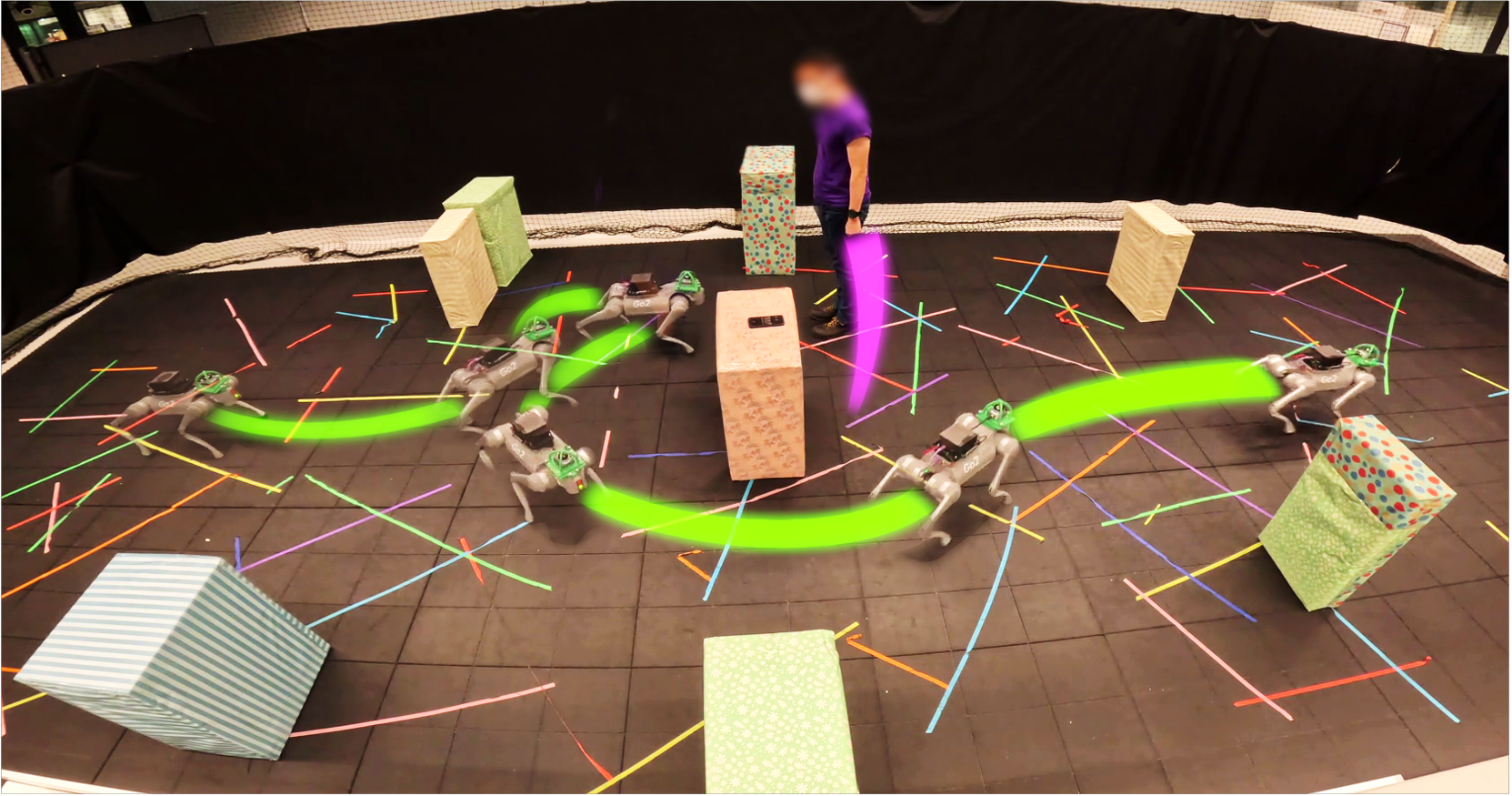}\\
			\subcap{(a) \ttt{Static}}  &\subcap{(b) \ttt{Dynamic}}\\
			\includegraphics[width=0.235\textwidth]{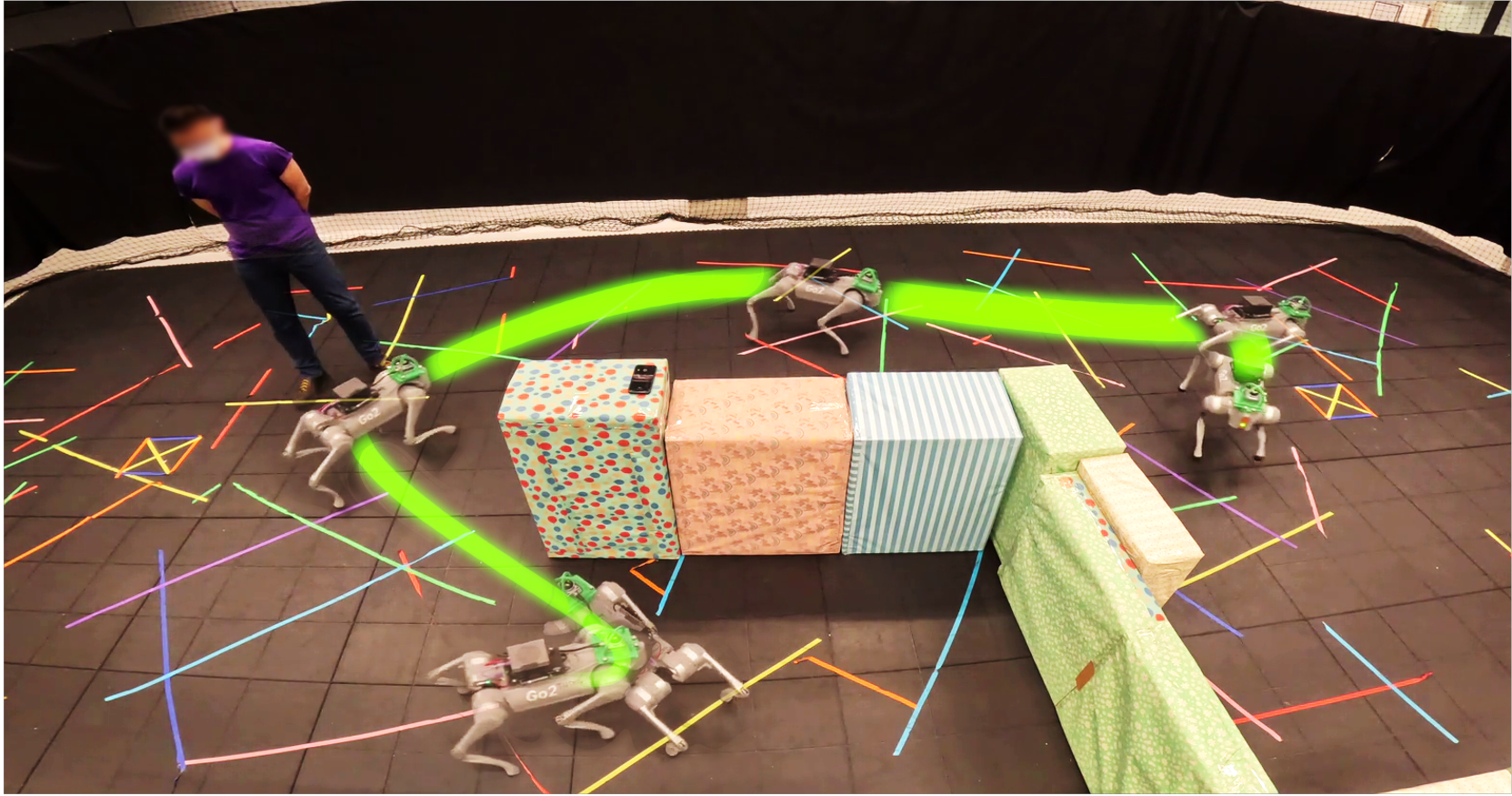}&
			\includegraphics[width=0.235\textwidth]{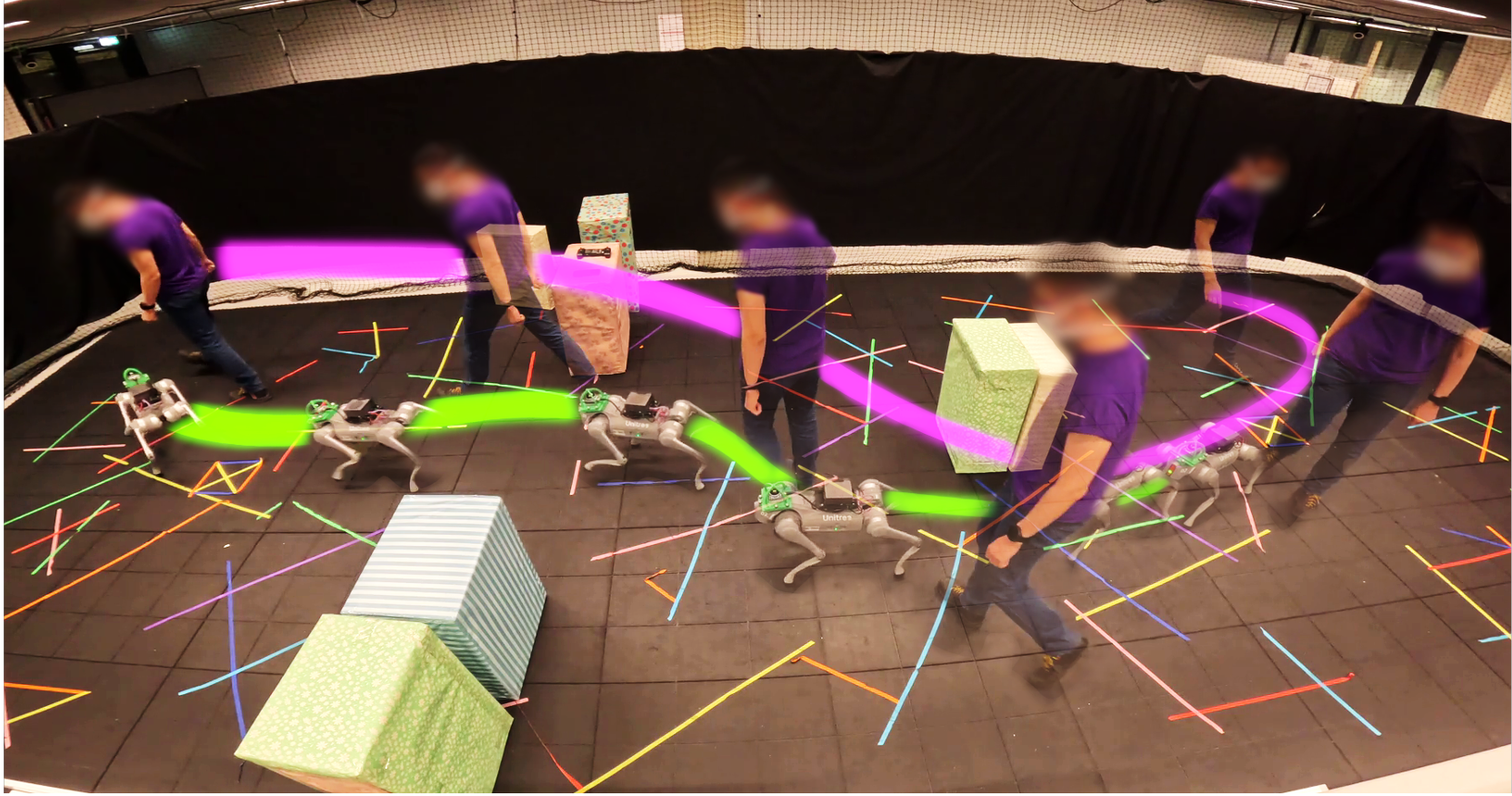}\\
			\subcap{(c) \ttt{DeadEnd}} &\subcap{(d) \ttt{Herding}}\\
		\end{tabular}
		\caption{Real-world, fully onboard tests for REASAN.}
		\label{fig:real}
		\vspace{-5mm}
	\end{figure}
	
	\begin{figure*}[t!]
		\vspace{1mm}
		\centering
		\includegraphics[width=\textwidth]{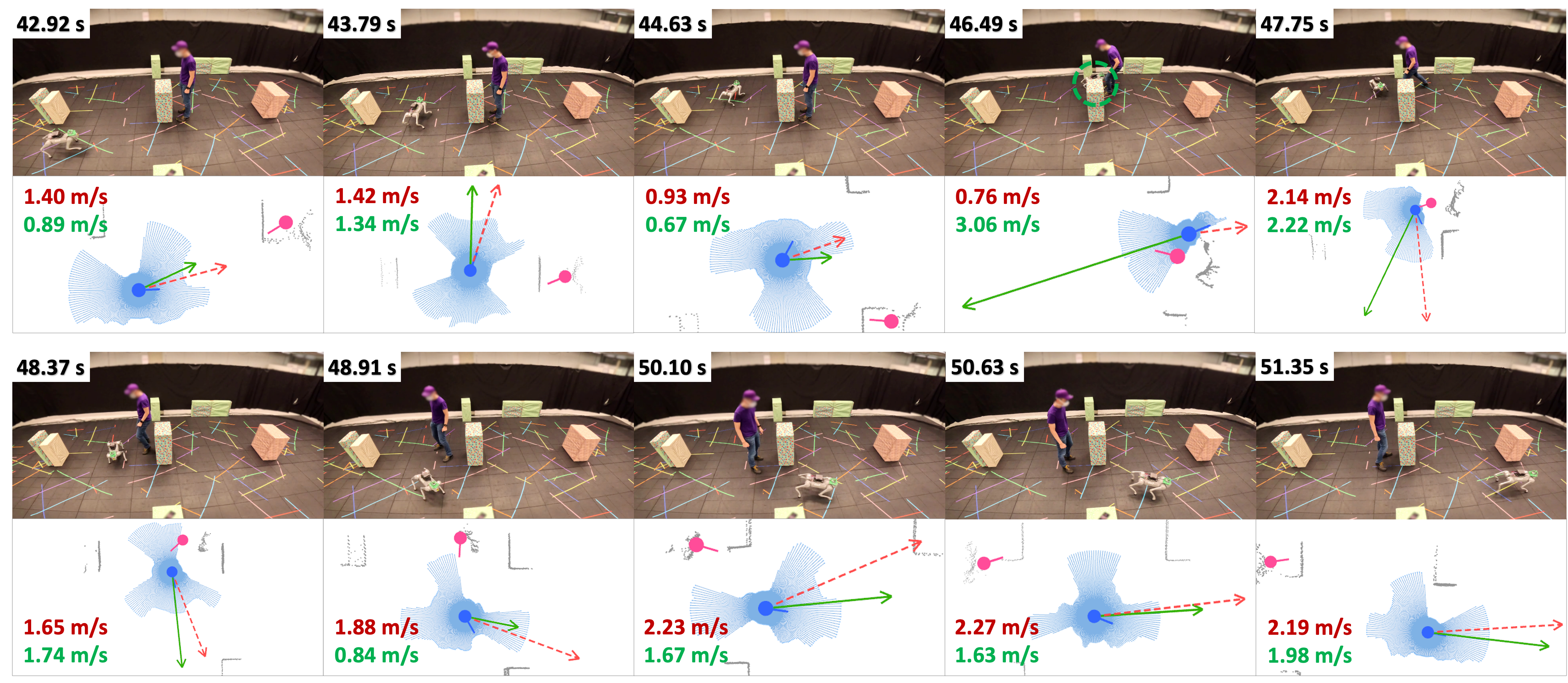}
		\captionof{figure}{Case study on dynamic obstacle avoidance. Dot in magenta depicts the dynamic obstacle position. Exteroceptive estimator output is depicted by blue rays given the raw point clouds in gray. Dotted red and green arrows are the navigation and safety-shield outputs, respectively, where the navigation command is reactively adjusted into a safe one given exteroception.}
		\label{fig:case1}
		\vspace{-3mm}
	\end{figure*}
	
	\subsection{Reactive Navigation}
	\subsubsection*{Robustness}
	We conduct experiments in four different scenarios within a \SI{40}{\meter\squared} test ground as shown in~\figref{fig:real}. To evaluate the full REASAN system, the robot is required to traverse between two locations $10$ times autonomously while handling various conditions: \ttt{Static}, involving complex scattered obstacles; \ttt{Dynamic}, with scattered static obstacles and a person who may suddenly block the pathway; and \ttt{DeadEnd}, where one end is set at a dead end with a person moving unpredictably nearby. Additionally, we introduce \ttt{Herding}, where only the safety-shield policy is activated with zero commanded velocity, and the robot is guided purely through collision-avoidance reactions as a person herds it between two ends for $5$ rounds. All tests are run with a speed limit of \SI{2}{\meter/\second}. Tab.~\ref{tab:test} summarizes each test's completion duration, and the robot exhibits no collisions with any obstacle, demonstrating REASAN’s robustness for safe reactive navigation in complex environments.
	
	\begin{table}[htbp]
		\centering
		\caption{Collision-free completion duration.}
		\setlength{\tabcolsep}{4pt} 
		\begin{tabular}{l|cccc}
			\toprule
			\textbf{Test Case} & \ttt{Static} & \ttt{Dynamic} &\ttt{DeadEnd} &\ttt{Herding} \\
			\midrule
			Time (\SI{}{\second}) &$270$  &$180$ &$190$ &$180$\\
			\bottomrule
		\end{tabular}
		\label{tab:test}
	\end{table}
	
	\subsubsection*{Modular reaction}
	We further investigate REASAN's obstacle-avoidance behavior across modules in a \ttt{Dynamic} scenario given in \figref{fig:case1}. When a sudden movement blocks the pathway, the robot exhibits a timely response in both the exteroceptive estimator and the safety-shield network, leading to a rapid correction of the velocity command that guides the robot to retreat and reactively select an alternative route. The safety-shield policy frequently outputs a lower speed than the navigation input due to nearby obstacles, showing its effectiveness in enforcing safe behavior. Moreover, the proposed exteroceptive estimator shows strong generalizability, effectively handling human obstacles despite their absence in the training data.
	
	\subsubsection*{Multi-robot reactive navigation}
\begin{figure}[t!]
		\centering
		\begin{tabular}{c}
			\includegraphics[width=0.41\textwidth]{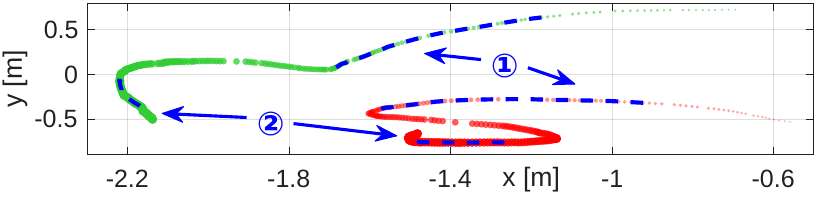}\\ 
			\hspace{1mm}\includegraphics[width=0.41\textwidth]{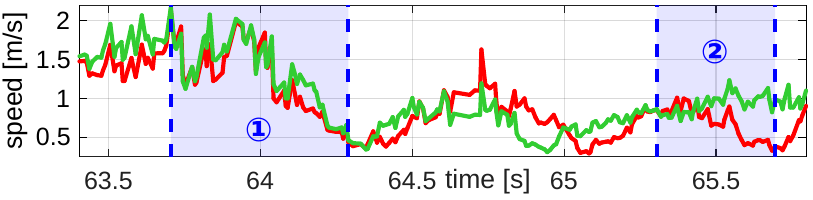}
		\end{tabular}
		\caption{Motion analysis in multi-robot tests. Blue curves and areas indicate reactive obstacle avoidance behavior.}
		\label{fig:multirobot}
	\end{figure}

	REASAN demonstrates successful sim-to-real transfer in a multi-robot setting, using identical hardware and the same deployment procedure described in \secref{subsec:deploy}, without any robot-specific tuning. We recreate a scenario similar to \figref{fig:real}-(a) with more free space, where two robots traverse between the two ends without any communication or coordination. This results in frequent, unpredictable, and highly dynamic interactions that require continuous reactive navigation, as illustrated in \figref{fig:front}. Despite these challenges, the experiment runs for \SI{93}{\second} without any collision. As shown in \figref{fig:multirobot}, quantitative insights are obtained using an onsite motion-capture system composed of $8$ Qualisys Miqus M3 cameras operating at \SI{100}{\Hz}. Within \SI{2}{\second}, the two robots perform mutual reactive avoidance twice while walking toward closely aligned goals, exhibiting abrupt speed variations between \SI{0.2}-\SI{2}{\meter/\second} within reaction distances of \SI{0.5}-\SI{1.2}{\meter}.
	
	\subsubsection*{Runtime efficiency}
\begin{figure}[t]
		\centering
		\setlength{\tabcolsep}{1pt} 
		\begin{tabular}{cc}
			\includegraphics[width=0.235\textwidth]{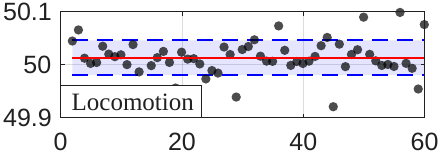} &
			\includegraphics[width=0.235\textwidth]{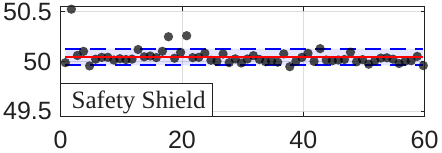}\\
			\includegraphics[width=0.235\textwidth]{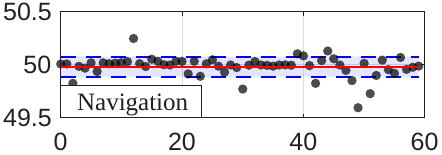} &
			\hspace{1.7mm}\includegraphics[width=0.222\textwidth]{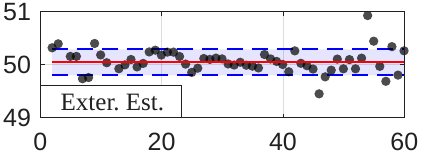}\\
		\end{tabular}
		\caption{Module runtime frequency (\SI{}{\hertz}). Red line and blue band depict the mean $\pm1$ standard deviation, respectively.}
		\label{fig:freq}
		\vspace{-2mm}
	\end{figure}

	We record onboard runtime frequencies for each network module in a \ttt{Dynamic} test. As shown in \figref{fig:freq}, all modules maintain stable real-time performance at \SI{50}{\hertz} over \SI{60}{\second}, as designated for deployment in \figref{fig:pipeline}. 
	
	\subsection{Limitations and Discussion}
\begin{figure}[t]
		\vspace{3mm}
		\centering
		\setlength{\tabcolsep}{0pt}
		\begin{tabular}{cc}
			\includegraphics[width=0.24\textwidth]{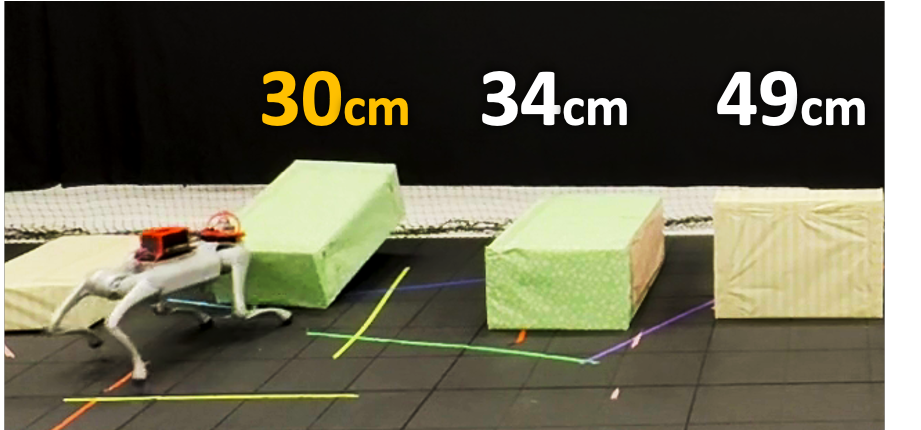}&
			\includegraphics[width=0.23\textwidth]{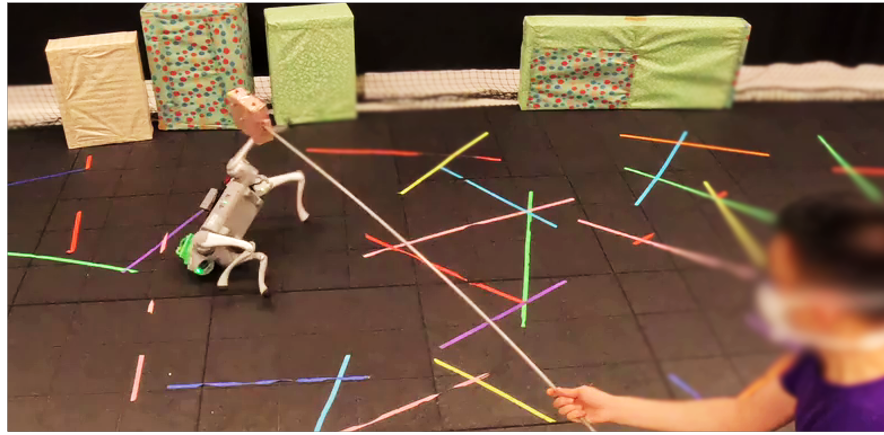}\\
			\subcap{(a) Low obstacle} & \subcap{(b) Rear-overhead attack}
		\end{tabular}
		\caption{Observed limitations of REASAN.}
		\label{fig:failure}
		\vspace{-4mm}
	\end{figure}

	REASAN demonstrates robust reactive obstacle avoidance and navigation in complex dynamic environments, running fully onboard quadrupedal robots in real time. Compared with state-of-the-art counterparts~\cite{he2024agile,wang2025omni}, the proposed modular decomposition of the complex legged motor-control problem enables learning with concise, lightweight network architectures and standard training pipelines, while delivering more strategic reactive behaviors, such as making detours and escaping dead ends. However, a few limits of REASAN are observed in our tests. The exteroceptive estimator is trained with obstacles of minimal \SI{0.5}{\meter}. Meanwhile, the LiDAR sensor is mounted on the front above the head with very limited capability to perceive nearby obstacles close to the ground. In reality, REASAN can avoid obstacles as low as \SI{0.34}{\meter}, showing generalizability to a certain extent, however, e.g., not \SI{0.3}{\meter}, which cannot be seen by the LiDAR (\figref{fig:failure}-(a)). This can be possibly addressed by enhancing the perception hardware by adding a LiDAR closer to the ground and following the training pipeline of the exteroceptive estimator in \secref{subsec:estimator}, whereas the policy networks can remain untouched. \figref{fig:failure}-(b) demonstrates another failure case where the robot falls over to the ground, when a small box hanging over a long rod is quickly approaching from the back over it. As the simulation does not include small overhanging obstacles and the LiDAR inherently struggles with high-speed perception, this scenario produces out-of-distribution inputs to the safety-shield network.
	
	\section{Conclusion}\label{sec:conclusion}
	REASAN is a modularized end-to-end reactive navigation framework for legged robots, comprising three RL-based policy networks for locomotion, safety shielding, and navigation, together with a Transformer-based exteroceptive estimator. Unlike existing monolithic end-to-end approaches that rely on heuristics or policy-switching mechanisms, our proposed modularization enables efficient training of lightweight neural networks using standard RL practices with more targeted reward shaping and curriculum learning. This design choice is validated by extensive simulation ablations. Real-world experiments demonstrate that REASAN achieves robust reactive navigation in complex environments fully onboard  in real time.
	For future work, we plan to develop more strategic reactive navigation policies through longer-horizon exteroceptive prediction. Extending the system to operate on uneven terrain is another promising direction. Moreover, incorporating cameras for semantic-aware perception could improve robustness in highly cluttered or unstructured environments where geometry alone is insufficient.
	
	\bibliographystyle{IEEEtran.bst}
	\bibliography{bibliography.bib}
	
\end{document}